# A Generic Model for Swarm Intelligence and Its Validations

WENPIN JIAO
Key Lab. of High Confidence Software Technologies (MOE)
Department of Computer Science and Technology
Peking University
No. 5, Yiheyuan Road, Beijing 100871
CHINA

*Abstract:* - The modeling of emergent swarm intelligence constitutes a major challenge and it has been tackled in a number of different ways. However, existing approaches fail to capture the nature of swarm intelligence and they are either too abstract for practical application or not generic enough to describe the various types of emergence phenomena. In this paper, a contradiction-centric model for swarm intelligence is proposed, in which individuals determine their behaviors based on their internal contradictions whilst they associate and interact to update their contradictions. The model hypothesizes that 1) the emergence of swarm intelligence is rooted in the development of individuals' internal contradictions and the interactions taking place between individuals and the environment, and 2) swarm intelligence is essentially a combinative reflection of the configurations of individuals' internal contradictions and the distributions of these contradictions across individuals. The model is formally described and five swarm intelligence systems are studied to illustrate its broad applicability. The studies confirm the generic character of the model and its effectiveness for describing the emergence of various kinds of swarm intelligence; and they also demonstrate that the model is straightforward to apply, without the need for complicated computations.

## 1 Introduction

Researchers have been inspired by many phenomena at the physical, chemical, biological, or social level to study the emergence of swarm intelligence [1][2][3][4][5][6][7]. Many models, mechanisms, and techniques for swarm intelligence have been proposed, such as the Particle Swarm Optimization (PSO) [8], Gravitational Search Algorithm (GSA) [9], Intelligent Water Drop (IWD) [10], Ant Colony Optimization (ACO) [11], Artificial Bee Colony (ABC) [12], and Holonic system model [13]. They have been applied in practice to enable the emergence of swarm intelligence of specific application systems (e.g., [14][15][16][17][18][19][20][21]).

However, those approaches (including models, mechanisms, and techniques) have been proposed for their respective application areas, rather than as general models. For an approach, its application swarms need to be in compliance with the specific requirements of the area, i.e., the swarms possess the features specified by the approach, so that the expected emergent properties can be exhibited. To apply a specific approach to a swarm, people need to be well aware of the behavioral features of the autonomous elements (or individuals) involved in the swarm and be convinced that the elements' behaviors accord with the approach. For instance, in order to apply the ant colony algorithm, the individuals in the swarm need to have the capability to behave like ants: they need to produce and spread pheromone-like information and further intensify their actions after perceiving the pheromone.

Furthermore, due to the unique characteristics of swarms in the real world, the features they possess may not fully comply with the requirements of any approach, so the expected swarm intelligence cannot be assured to emerge even though the most appropriate approach is selected and applied. To implement a system with swarm intelligence, people have to repeatedly try different approaches in order to discover the most suitable one. This leads to a high computational complexity and development cost even without a guarantee of the expected swarm intelligence emerging.

Therefore, if we can discover and build a generic model (or mechanism) that is applicable to any type of swarm intelligence, the application of swarm intelligence to diverse problems can be greatly simplified and the expected swarm intelligence can be more likely to successfully emerge. More specifically, the research question investigated here is as fol-lows: is there a generic model for the emergence of swarm intelligence?
In this paper, we propose a generic model for the emergence of swarm intelligence. To achieve this, we first analyze the intrinsic essence that drives the development and emergence of swarm intelligence in Section 2. In Section 3, we compare our model to established existing approaches. In Section 4, we put forward a generic model. To illustrate that the proposed model is broadly applicable, we present five swarm intelligence systems in Section 5. In Section 6, we discuss and validate our model by experiments. In Section 7, we make some concluding remarks and point out the direction of our future work.

# 2 Background and Motivation: Intrinsic Driving Forces of Swarm Intelligence

## 2.1 Established View

In the existing approaches in the literature, it is generally understood that the exposed appearances (including static properties and dynamic behaviors) of individuals involved in a swarm depend on how individuals react to changes occurring in the external environment (including other situated individuals). For instance, in the Ant Colony Algorithm [11], ants decide or choose their actions according to the pheromones secreted in the environment; while for birds (or particles) involved in the Particle Swarm Optimization Algorithm [8], they change their flying velocities and directions according to the states of others surrounding them. Existing approaches usually hypothesize that the emergence of swarm intelligence is rooted in individuals' reactions to the environment [22][23]. Therefore, the research has consistently focused on how the appearances of individuals are impacted by the environment when implementing swarm intelligence.
However, environmental factors are not the necessary conditions (or inevitable factors) that individuals use to decide their actions. For instance, when the outside temperature drops and the weather gets cold, some people may put on more clothes whilst others may not. Whether someone wears additional clothes depends on whether or not they feel cold. An individual may not feel cold even if the weather turns cold, i.e., the change in weather does not determine an individual's decision on clothing. The weather's change is just an external condition that may influence the decision, and the influence may take effect only when the weather's change incidentally makes a person feel cold.

## 2.1 Dialectical Contradiction Based View

Essentially, environmental factors exist and these conditions are present in the external periphery environment of individuals; they are usually referred to as external causes in dialectics. However, existing approaches falsely presume that the external causes determine the emergence of swarm intelligence, and as a result they are unable to model swarm intelligence precisely. In addition, as the complexity of the environments increases, external causes are usually innumerable and unpredictable. We cannot foresee and capture all of the relationships between the appearances of individuals and the environmental factors, let alone completely anticipate and specify all of the reactions of individuals to the environmental changes. This contributes to the challenge of adapting the existing swarm intelligence approaches to complex systems situated in the real world.
In dialectics, external causes are the outside conditions that influence the development of an entity whilst internal causes are the inner sources of the development. The external causes can only take effect through internal causes. For instance, when the weather gets cold (external cause), this can make people feel cold (internal cause) and may trigger the behavior of putting on more clothes.
Internal causes are essentially inner contradictions of entities (here a contradiction is a concept in the philosophical sense and it refers to a dialectical contradiction instead of a logical contradiction). A contradiction is a unity of opposites (opposites, in turn, are the two aspects of a contradiction) [24]. For example, coldness and hotness are the two aspects of a contradiction. A contradiction defines a relation between two attributes (i.e., two aspects) of an entity. The relation reflects both unity and struggle, i.e., the two aspects are mutually exclusive while at the same time they are reciprocal.
It is believed that contradictions are the intrinsic sources and forces driving the development and change of entities [24]. From this perspective, contradictions exist in the process of all entities, i.e., contradictions are inevitable and the development of the world does not occur without them. In addition, all entities exist within the unities of contradictions; their properties and development are determined and driven forward by contradictions. In the case of biological organisms, their properties are determined by internal genes that occur in allelic genes. Alleles are like contradictions, and they control the characteristics of organisms.
As the external causes that influence the appearances of entities, the environment acts as the situation and medium for entities' interactions to exert influ-ences on entities' contradictions. Accordingly, entities are triggered to change their behaviors and exhibit varied appearances. For instance, by heating or cooling, the environment can influence the feelings of people on hotness or coldness and subsequently trigger people's behavior of adding or reducing clothes.

Within the context of a swarm, all of the individuals are associated by direct (or indirect) interactions through the environment; these interactions can be viewed as influencing and altering the contradictions of the individuals. For any individual, the others (except for itself) in the swarm are also part of the individual's environment, so their interactions can naturally be considered to occur between the individual and its environment. When interacting individuals present a special status (or property) as a whole, we can assert that some swarm intelligence emerges.

To the best of our knowledge, a contradiction based swarm intelligence model has not been proposed in the literature.

Therefore, we propose a generic model based on contradictions for the emergence (or formation) of swarm intelligence. In the model, firstly, the contradictions of individuals determine the appearances of individuals; secondly, individuals involved in the swarm are related and interact through the environment; thirdly, the appearances of individuals restrict the relevances and interactions of individuals; fourthly, interactions among individuals influence the contradictions of individuals; and finally, swarm intelligence emerges from the overall development of the contradictions of individuals involved in the swarm.

# 3 Related Work

The investigation of the emergence phenomena of swarm intelligence has attracted substantial attention, and many principles and models had been proposed. Among those principles and models, some are concerned with and inspired by specific emergence phenomena [25], whereas others try to discover and establish the foundation of emergence.

## 3.1 Approaches Inspired by Specific Emergence Phenomena

Nowadays, many computational algorithms or models have been presented for swarm intelligence [26][27][23][28][29][30] and also successfully applied in a variety of problem domains [31][41][32][33][34]. Most of these approaches are inspired by natural or social phenomena. Some of the most representative and significant ones include Deneubourg's cellular automata model (CA) devoted to cemetery formations in ant colonies [14], Kennedy's particle swarm optimization algorithm (PSO) [8], Dorigo's ant colony optimization algorithm (ACO) [11], and Karaboga's artificial bee colony (ABC) [12].

CA [35] imitates the self-reproductions of living systems, in which cells (i.e., individuals) scatter in a lattice and follow the same rules of actions; the evolution of the system results from the interactions of a large number of cells. Although CA is applied widely, for instance, to simulate the formation of an ant cemetery [36][37], CA is quite difficult to apply to emergence phenomena taking place in the real world because the cellular space and evolution rules are so strict.

ACO is inspired by the behavior of foraging ants [11]. In the ant colony, through the positive feedback of secreting pheromones, ants (or individuals) can quickly search for the best path to food. ACO requires that the search space and the transitions between states can be accurately represented, the qualities of solutions can be defined and evaluated, and there is a positive feedback mechanism for updating pheromone concentrations.

PSO simulates the behavior of birds (i.e., particles) searching for food [8]. In the swarm, particles update their velocities and positions to find the best solution in the search space in a collaborative manner. This is based on the experiences of a particle and the experiences of its surrounding particles. PSO has the advantages of simplicity, fast convergence, and the need for few parameters, but it does not work well if particles' movements cannot be defined clearly and uniquely [38].

ABC mimics the honey collection behavior of a bee colony [12]. In a colony, bees have a clear division of labor and they conduct their activities according to their roles; the bees share information about the nectar source in the colony. ABC synthesizes global and local searches to find the optimal solution (i.e., the largest nectar source). However, in ABC, the roles of bees are pre-appointed and unchangeable.

In summary, however, these existing approaches do not embody the nature of swarm intelligence and they are not generic enough to describe many types of emergence phenomena of swarm intelligence. They usually have many assumptions which bring great difficulty for researchers in adapting them appropriately to concrete application problems [39]. For instance, assumptions include that the individuals have similar or the same characteristics (e.g., behavior rules and ways of interacting with others and the environment), the individuals have pre-appointed fixed roles, and the specific rules for the evolution of the swarm intelligence are pre-specified and applied. Table 1 presents four basic properties of the representative approaches: ways of driving or selecting individuals' behavior; evolution of the swarm intelligence; main areas of application; and limitations or assumptions.

**Table 1. Properties of Swarm Intelligence Models**

| Model | Way of Behavior | Evolution of Swarm | Areas of Application | Limitation/Assumption |
|---|---|---|---|---|
| CA | Driven by space states, fixed and same action rules | Survival and Reproduction | Formation, Classification | Limited and regulated space, and strict evolution rules |
| ACO | Stimulated by pheromone concentrations, and positive feedback | Good solution spread | Routing, Scheduling, Combination Optimization, Task Assignment | Accurate transitions, and evaluable solutions |
| PSO | The tradeoff between learning from experiences and ensuring safety | Good solution diffusion | Global Optimization, Classification, Clustering, Scheduling | The clear way to uniquely define better positions, and pre-designed roles |
| ABC | Fixed division of labor and behavior pattern | Good information propagation | Collaborative Problem Solving, Clustering, Classification, Task Assignment | Pre-assigned roles, and specified behavior pattern correspondingly |

Compared with these existing approaches, our model does not impose any limitations on individuals or the environment. In our model, neither the roles nor the behavior patterns (or action rules) of individuals need to be pre-defined. Individuals decide their actions completely based on their internal contradictions. When individuals make decisions, they need not pay close attention to the environment (including other individuals) unless the interactions with the environment result in updates of their internal contradictions. Individuals' roles are determined dynamically when their internal contradictions are updated, and their behavior patterns change correspondingly when they play different roles.

## 3.2 Generic Principles or Models

Because emergence is usually considered as a fundamental property of self-organizing systems, many research results related to the emergence of swarm intelligence appear in the self-organization domain [40][41]. For instance, Ashby's principle of self-organizing dynamic systems based on attractors [42], Von Foerster's principle of order out of noise based on random fluctuations [43], Prigogine's entropy production-minimum principle [44], and Haken's slaving principle [45] based on far-from-equilibrium.

Besides these general principles for explaining the emergence of swarm intelligence, researchers have also endeavored to establish generic or unified models for swarm intelligence.

In [39], a general algorithmic structure of intelligent optimization systems is proposed, which combines existing optimization algorithms by designing a modular framework and modularizing the features and functions of those algorithms. This work tries to integrate all algorithms (or techniques) into a unified framework instead of building a new model. Similarly, in [46], an integrated optimization approach is described based on a meta-modelling technique. As a consequence, the proposed frameworks cannot avoid the limitations of the existing approaches and its application areas are also restricted to specific problems they address.

In [23], the author tries to discover and explain the essential features of the emergence of swarm intelligence. In his opinion, swarm intelligence emerges from interactions among individuals under the constraints that the environment imposes, and it evolves in response to the dynamics of the environment. The author believes that swarm intelligence is rooted in the constraints and dynamics of the environment. With a similar belief, the authors in [22] argue that emergence results from fine-tuning or adapting the constraints imposed by the logical or physical environment; they propose a theory to assess their argument. However, as discussed previously and as shown in our model, swarm intelligence is a reflection of the distribution of internal contradictions of individuals, and the emergence and evolution of swarm intelligence is rooted in the development of internal contradictions in individuals.

Although the principles mentioned above have been successfully applied to many domains and described various natural phenomena, they are too general to be practical. At the other extreme, specific approaches inspired by natural or social systems have been proposed that encompass strict limitations and can only be applied to specific applications. In contrast, our proposed model is both general and practical. Every individual has its contradictions, which are independent of a specific environment, so the behaviors and interactions of individuals can be specified and implemented independently of any environments; this exhibits the generality of our model. In addition, the contradictions of individuals can be easily analyzed, defined, and implemented with low computational complexity; this shows the practicability of our model. Our model is not specific to any particular swarm intelligence and can be easily applied to various emergence phenomena. It effectively bridges the gap between general principles and specific approaches.

# 4 Swarm Intelligence Model based on Contradiction

In this section, we present a novel formation model for swarm intelligence, including how individuals involved in the swarm are driven by contradictions to determine their appearances (particularly their behaviors), how individuals are associated and interact to influence their contradictions, and how swarm intelligence emerges from the development of individuals' contradictions.

## 4.1 Formation Model

A swarm is situated in the environment (which *mediates* the behaviors and interactions of individuals involved in the swarm), and it can be divided into two levels (Fig.1): the individual level and the swarm level.

At the individual level, an individual is abstracted as the unity of contradictions and appearances, i.e., an individual *consists* of contradictions and *presents* appearances and the contradictions *determine* the appearances of the individual. At the swarm level, the swarm is abstracted as the relevances and interactions of individuals, from which the swarm intelligence emerges. The individual level and the swarm level are interdependent. On one side, the appearances of individuals contain the swarm intelligence and conversely, the swarm intelligence is the overall reflection of the appearances of individuals. On the other side, the swarm is the *situation* of relevances and interactions of individuals, that is, individuals are involved in the swarm and they interact when associated.

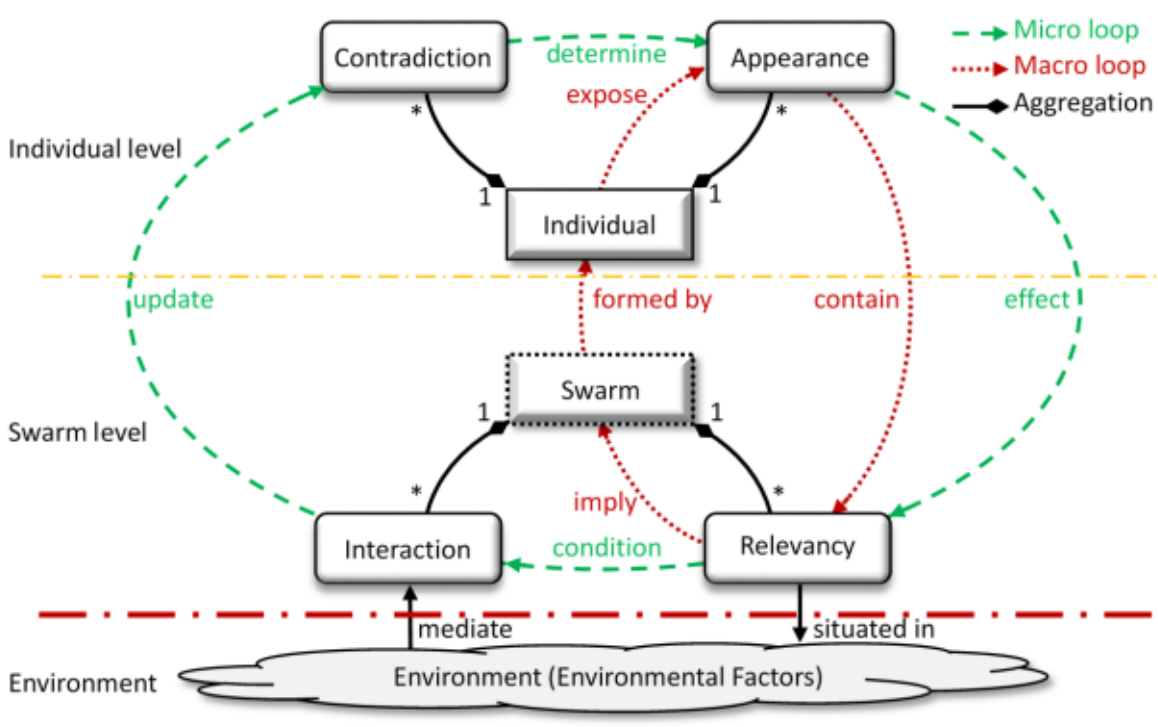

**Fig.1 Formation Model of Swarm Intelligence**

In a swarm, there are two control loops. One is the micro contradiction development loop, i.e., "*Contradiction* → *Appearance* → *Relevance* → *Interaction* → *Contradiction*" loop (see the outside loop in Fig.1), and the other is the macro swarm intelligence emergence loop, i.e., "*Swarm* → *Individual* → *Appearance* → *Relevance* → *Swarm* (*Intelligence*)" loop (see the inside loop in Fig.1). In the micro loop, individuals' contradictions *determine* their appearances, their appearances (particularly the exposed behaviors) influence (or *effect*) their relevances in the swarm, their relevances *condition* their interactions, and their interactions *update* their contradictions. In the macro loop, the swarm is *formed* by individuals, individuals' appearances *contain* their relevances, and the relevances *imply* the swarm intelligence.

## 4.2 Individual

An *individual* contains a collection of contradictions and presents a series of appearances. These appearances are determined by the contradictions.

$$ind = <\Gamma, A, \delta>$$

where, $\Gamma$ is the set of inherent contradictions, $A$ is the set of outward appearances, and $\delta$ is the set of determination functions defining the determination relationships between the contradictions and appearances.

- *Contradiction*. A contradiction consists of two contradictory aspects, i.e., one aspect and its opposite.

$$c_{\in\Gamma} = <o, \bar{o}, \varsigma>$$

where, $o$ and $\bar{o}$ are the opposite aspects, i.e., positive and negative aspects respectively, and the positive and the negative are relative, i.e., $\bar{\bar{o}} = o$; $\varsigma$ is the relative strength of the contradiction, $\varsigma \in [-1, 1]$, and it reflects the balance of the strength of the two aspects, i.e., $\varsigma > 0$ $(or\ \varsigma < 0)$ if the strength of $o$ is bigger (or less) than that of $\bar{o}$. The bigger $|\varsigma|$ is, the more prominent the contradiction is.

- *Appearance*. An appearance is a property or behavior that the individual shows, for instance, speed, height, weight, color, advance/retreat, etc.

$$a_{\in A} = <\rho, e>$$

where, $\rho$ is the property or behavior embodied in the appearance, and $e$ is the effect(s) of the appearance. For a property, the appearance may have no effect, whilst for a behavior, the appearance may update the $\varsigma$ values of the contradictions of the individual and/or affect the environment.

- *Determination Function*. For each appearance, there is a corresponding determination function to calculate the visibility of the appearance.

$$d_{\in\delta}: 2^{\Gamma} \times A \to R$$

where, $R$ is the set of real numbers. Suppose $C \subseteq \Gamma$ $(C = \{c_1, c_2, \dots, c_n\})$ is the set of contradictions that dominates the visibility of an appearance. Let $W = \{w_1, w_2, \dots, w_n\}$ be the weights reflecting the im-portance of the contradictions in dominating the visibility; $w_i \in [-1,1]$ and $\sum_{i=1}^{n} |w_i| = 1$. $w_i > 0$ (or $w_i \leq 0$) implies that the contradiction $c_i$ may promote (or hinder) the appearance's visibility. In many cases, the visibility of an appearance can be simply determined by the weighted sum of the relative strength of those contradictions dominating the appearance, i.e., $d(a) = \sum_{i=1}^{n}(\varsigma_i \times w_i)$. An appearance is visible if $d(a) > 0$.

In addition, when an appearance manifests as a behavior, the visibility of the appearance reflects the probability of performing the behavior; when the appearance manifests as a property, the visibility reflects the probability of the property appearing.

## 4.2 Swarm

A swarm is formed by a group of associated and interacting individuals.

$$sw = < \Sigma, \Pi, X >$$

where, $\Sigma$ is the set of individuals, $\Pi$ is the set of relevances specifying the associations among individuals, and $X$ is the set of interactions among the associated individuals.

- *Relevance*. A relevance reflects the association between an individual and its surroundings (including other individuals and the *environment*). Relevances can occur between two or more individuals, and they form a complex association network. Nevertheless, we can simplify relevances into a basic form. In each basic relevance, there is a central individual and other individuals are associated with it via the environment.

$$r_{\in \Pi} = < \iota_c, I, E >$$

where, $I \subseteq \Sigma$ is the set of individuals involved in the relevance, $\iota_c \in I$ is the central individual, $E$ is the set of environmental factors, and all of the other individuals in $I$ are associated with $\iota_c$ through the specified environmental factors.

For every individual in a swarm, there is at least one relevance in which the individual is the central individual. Then, the complex association network in the swarm is the composite of these basic relevances.

- *Interaction*. Associated individuals interact with one another, and the interactions may result in 1) some contradictions being updated, 2) old contradictions disappearing and new contradictions arising within some individuals, or 3) new individuals being generated.

For the sake of simplicity, we assume that 1) all contradictions (including newly produced contradictions) have been pre-specified in the contradiction sets of individuals, 2) all individuals have been pre-specified, including a pool of new individuals that may be generated during interactions, and 3) new generated individuals are always the central individuals of some relevances at the time when they are generated. Thus, an interaction can be simply regarded as a process of changing the central individual's contradictions.

$$x_{\in X}: \Pi \rightarrow \Gamma$$

This indicates that an interaction always takes place in a relevance and updates the contradictions of the central individual. Of course, an interaction may result in contradiction updates of multiple involved individuals. In that case, we can specify multiple interactions specific to the involved individuals and then appoint each individual as the center of a single interaction.

## 4.3 Formation of swarm intelligence

Swarm intelligence is reflected in the overall performance of individuals, and it is often manifested as the emergence of some holistic structures (or organizations, e.g., migrating wild geese and bee colony) or global behavior modes (or patterns, e.g., ants' foraging and fish school's hedging). In essence, the emergence of swarm intelligence is the result of the continuous development of the individuals' contradictions within the swarm.

Inside an individual, some contradictions may become more prominent (i.e., the relative strength becomes stronger) whilst others become more inconspicuous (i.e., the relative strength becomes weaker) due to the behaviors of the individual or the interactions involving the individual. Gradually, the configuration of the contradictions may become stable and a specific configuration of the contradictions leads the individual to present an outward performance that is distinct from the other individuals (e.g., acting as the leading goose in migrating wild geese or the queen in the bee colony). Consequently, the positions (or roles) of individuals in the swarm are differentiated. When individuals occupying different positions (or playing different roles) are associated with one another in the swarm, they collectively form a specific society (or organization).

On the other hand, because individuals play different roles in the swarm, the same type of contradictions inside individuals are updated differently in interactions and subsequently, they present a distinct distribution among individuals. As a result, the appearances of individuals display some specific global modes (or patterns) in the swarm. For instance, in peas, for the contradiction related to the length of the stem (suppose the two aspects of the contradiction are "*long-stem*" and "*short-stem*", respectively), the aspect of "*long-stem*" dominates in three-fourths of peas, so three-quarters of the peas have with long stems. For another example, in the foraging ants, the contradictions related to the moving mode inside many ants are affected by the pheromones secreted in the environment (see Section 5), so there are always many ants that are moving along with a transportation route connecting the nest and the food source.

Essentially, the configuration of contradictions within an individual shows the horizontal connection among contradictions inside the individual; whilst the distribution of distinct types of contradictions in the swarm show the vertical connection among the same type of contradictions in the range of the swarm. The horizontal connections among contradictions determine the specializations of individuals and further the social positions of individuals, which imply the global organization of the swarm; whereas the vertical connections determine the distributions of individuals' properties and behaviors, which imply the global properties or behavior modes occurring in the swarm.

Suppose $N$ is the number of individuals in the swarm, i.e., $\Sigma = \{\iota_1, \iota_2, \dots, \iota_n\}$, and $M$ is the number of contradictions in an individual, i.e., $\iota_i.\Gamma = \{c_1, c_2, \dots, c_m\}$. Swarm intelligence can be described as a synthesis of the horizontal and vertical connections of contradictions.

$$
\begin{aligned}
&Intelligence_{sw} \\
&= Syn(\, H_{\iota_1}(c_1, c_2, \dots, c_m), \dots, H_{\iota_n}(c_1, c_2, \dots, c_m), \\
&\qquad V_{c_1}(\iota_1, \iota_2, \dots, \iota_n), \dots, V_{c_m}(\iota_1, \iota_2, \dots, \iota_n)\,)
\end{aligned}
$$

where, $H_{\iota_i}(c_1, c_2, \dots, c_m)$ is a horizontal connection function that specifies the configuration of contradictions within individual $\iota_i$, $V_{c_i}(\iota_1, \iota_2, \dots, \iota_n)$ is a vertical connection function that specifies the distribution of the contradiction $c_i$ in the swarm, and $Syn$ is the synthesizer function of the composition of $H_{\iota_i}$s and $V_{c_i}$s.

### 4.4 Summary

Briefly, the proposed model is based on two key points: 1) the emergence of swarm intelligence is rooted in the development of individuals' internal contradictions and the interactions between individuals, and 2) swarm intelligence is essentially a combinative reflection of the configurations of contradictions inside individuals and the distributions of contradictions over the swarm. Accordingly, the model divides a swarm into two levels. The first defines how individuals' appearances are determined by their internal contradictions and the second level describes how individuals are associated with one another as well as with the environment. There are two control loops involved in the model. The first is the micro contradiction development loop and the second is the macro swarm intelligence emergence loop. These loops interweave the two levels together and control the development of individuals and the swarm. In the micro loop, an individual's contradictions dominate its outward appearances (i.e., properties and behaviors) in a swarm; meanwhile, the contradictions are updated while an individual is associated and interacts with others in the environment. In the macro loop, swarm intelligence emerges from the associated and interacting individuals, and swarm intelligence arises when the contradictions of individuals present some distinct configurations and distributions as a whole.

## 5 Case Studies

To verify the effectiveness and generality of the proposed formation model for swarm intelligence, we implement five systems in different fields. Among these systems, some display a holistic structure (or organization), some display a global behavior mode (or pattern), and some display both.

### 5.1 Foraging Ants

In a foraging ant colony, idle ants search for a food source(s) and loaded ants transport food back to the nest. While loaded ants are moving back, they leave signs (i.e., pheromones) along the paths for themselves or others to find shortcuts to the food source(s) later. A foraging ant contains the two contradictions:

- $c_{11}$. Idleness *vs* Loadedness: Is the ant idle or busy transporting food (or loaded)?
- $c_{12}$. Location's ordinariness *vs* specialness: Is the ant at an ordinary place, i.e., there is nothing around, or at a special place, e.g., there is pheromone, a food source or the nest?

An ant shows a collection of appearances and these appearances are manifested as behaviors.

- $a_{11}$. Move randomly, if the ant is ***idle*** and at an ***ordinary*** place (i.e., $c_{11}.\varsigma = 1 \wedge c_{12}.\varsigma = 1$).
- $a_{12}$. Move toward the food source, along the path where the pheromone is deposited if the ant is ***idle*** and perceives there is ***pheromone*** *around* (i.e., $c_{11}.\varsigma = 1 \wedge c_{12}.\varsigma = pher$).
- $a_{13}$. Move backward to the nest and secrete the pheromone, to transport food back to the nest and secrete the pheromone along the way if the ant ***loads*** food and has ***not*** been back the ***nest*** (i.e., $c_{11}.\varsigma = -1 \wedge c_{12}.\varsigma < 0 \wedge c_{12}.\varsigma \neq nest$).
- $a_{14}$. Load food, if the ant is ***idle*** and at the site of the ***food source*** (i.e., $c_{11}.\varsigma = 1 \wedge c_{12}.\varsigma = food$).
- $a_{15}$. Unload food, if the ant is ***loaded*** and at the ***nest*** (i.e., $c_{11}.\varsigma = -1 \wedge c_{12}.\varsigma = nest$).

At the swarm level, every ant acts independently; it is connected to and interacts with the environment. The relevances of ants involve a series of environmental factors, such as the pheromones, the location of the nest, and the site of the food source. Within an interaction, the central ant's contradiction of $c_{12}$ is updated according to the place where the central ant is located.
The behavior of foraging ants is displayed in Fig.2, in which the brown area on the left side denotes the nest, the red area on the right represents the food source.

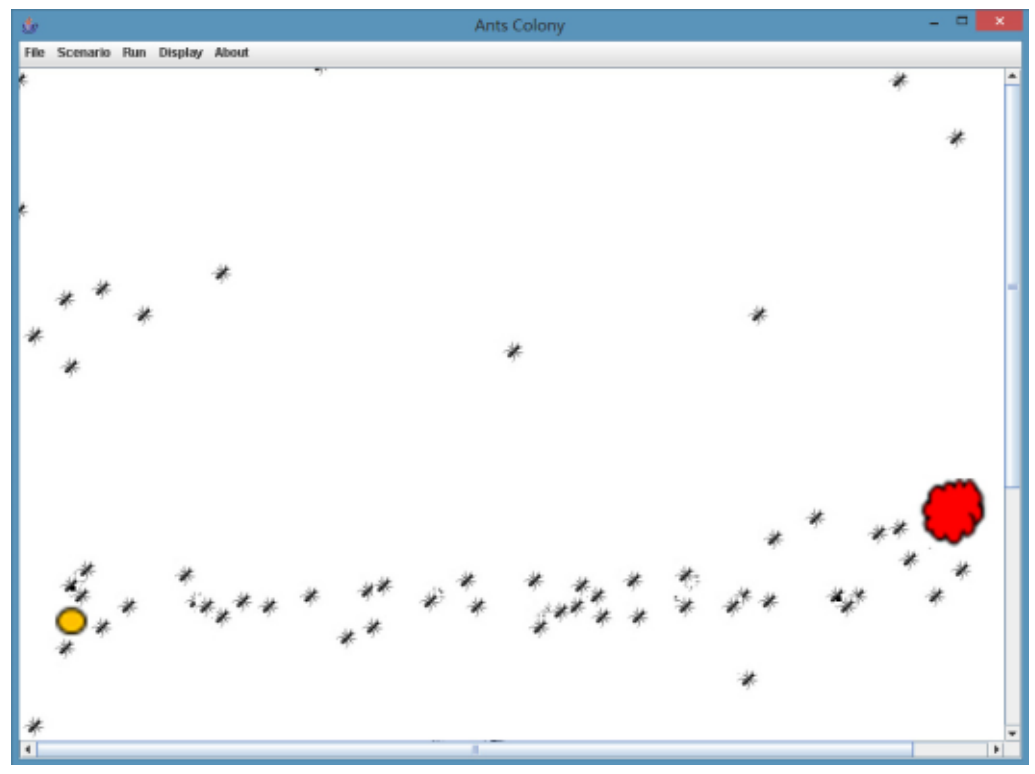

**Fig.2 Foraging Ants**

The swarm intelligence of ants is the emergence of a specific behavior pattern. Correspondingly, the movements of the ants forge an emergent path between the nest and the food source.

## 5.2 Development of Queen Bee

In the bee colony, every bee has a special endocrine gland that can secrete a kind of pheromone (referred to as *queening pheromone* below). This pheromone can promote the development of the endocrine gland and the bee itself, but also inhibit the development of the endocrine glands of the surrounding bees.
Inherently, a bee contains one contradiction related to the queening development (essentially related to the endocrine gland that secretes the queening pheromone):

- $c_{21}$. Endocrine gland's Development *vs* Degradation: is the endocrine gland developing or degrading?

The development and degradation of a bee's endocrine gland determines whether the bee has the potential of becoming a queen or not. When the relative strength of the contradiction is biased toward development (i.e., $c_{21}.\varsigma > 0$), the endocrine gland becomes more mature and the bee can potentially become a queen. A bee becomes a queen when its endocrine gland has matured. On the contrary, when the relative strength of the contradiction is negative, the endocrine gland degrades gradually; when the gland ultimately dysfunctions (i.e., $c_{21}.\varsigma \approx -1$), the bee loses the ability to compete for queening completely.
A bee shows the behavior of secreting the queening pheromone. Obviously, the objective of executing this behavior is to boost the bee's competitiveness for queening (i.e., to make the relative strength of the contradiction of $c_{21}$ bigger) and to damage the competitiveness of others. In addition, along with the maturing of the endocrine gland, the bee has a greater potential of becoming the queen and present the property of queening.
At the bee colony level, whenever a bee encounters another bee, it secretes the queening pheromone to promote its own endocrine gland's development and simultaneously inhibit the opposite's endocrine gland's development (as a result, the opposite's endocrine gland degrades gradually). The strength of promotion (or inhibition) is determined by the amount of the pheromone, and the amount of the pheromone depends on the maturity of the endocrine gland. Generally, the more mature the endocrine gland is, the larger the amount of the pheromone that is secreted by the gland. That is to say, a bee is associated and interacting with another bee by the pheromone secreted by both of the two bees. Through the interaction between the two bees, the relative strength of one bee's contradiction of $c_{21}$ is strengthened whilst the relative strength of the other bee's contradiction is weakened.
In the queen bee's development, the population size of the bee colony is 2000, the maturity degrees of their endocrine glands are randomly distributed (approximately 20% of the population have the potential of being queen, i.e., $c_{21}.\varsigma > 0$), and a bee encounters other bees randomly and interacts with them mutually. In the implementation, a bee interacts with less than 100 bees, instead of the whole swarm, in each encounter.
It shows that the number of bees that are potentially queening (i.e., $c_{21}.\varsigma > 0$) decreases quickly; after 48 days, there is only one bee winning the queening competition (Point *a* in Fig.3). Furthermore, when the old queen bee dies, another queen bee appears in 15 days (Point *b* in Fig.3).

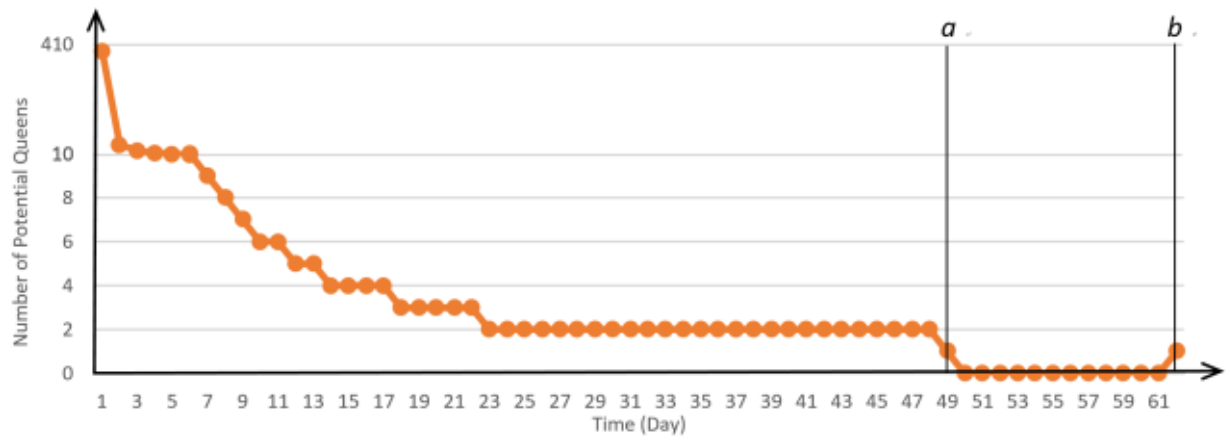

**Fig.3 Number of Potential Queen Bees**

The swarm intelligence emerging from the bee colony manifests as a distinct global organization, i.e., there is at most one queen bee in the bee colony.

## 5.3 Migrating Wild Geese

In a group of migrating wild geese in flight, every goose keeps a safe distance from the others to avoid collisions while following its preceding geese as closely as possible to save its strength; it can fly away from others (to be *stray*) so that it could fly more freely or remaining close to others (to be *gregarious*) so that it could be well protected by the group. When a goose cannot see the leading goose, i.e., the leading goose is out of the range of vision, the goose strays and loses the protection of the group. A migrating wild goose contains four contradictions:

- $c_{31}$. Leading *vs* Following: Is the goose leading the group or following others?
- $c_{32}$. Vigorousness *vs* Tiredness: Is the (leading) goose energetic or tired?
- $c_{33}$. Energy-savingness *vs* Distance-safety: When the goose follows another goose, can it save more energy by flying closer to the preceding goose or is the following distance too close to be safe?
- $c_{34}$. Straying *vs* Gregariousness: Is the goose stray so that it could fly freely or is it gregarious so that it could be well protected by the group?

The leading goose is responsible for leading the group unless it is tired. A follower goose tries to save energy while keeping a safe distance from the preceding goose; meanwhile, it constantly watches the leading goose to assure itself that it does not lose the protection from the group while pursuing adequate flying freedom. That is to say, a follower always tries to maintain a balance between energy-savingness and distance-safety and between straying and gregariousness as well; this keeps the relative strengths of both the contradictions of $c_{33}$ and $c_{34}$ to tend to be 0.

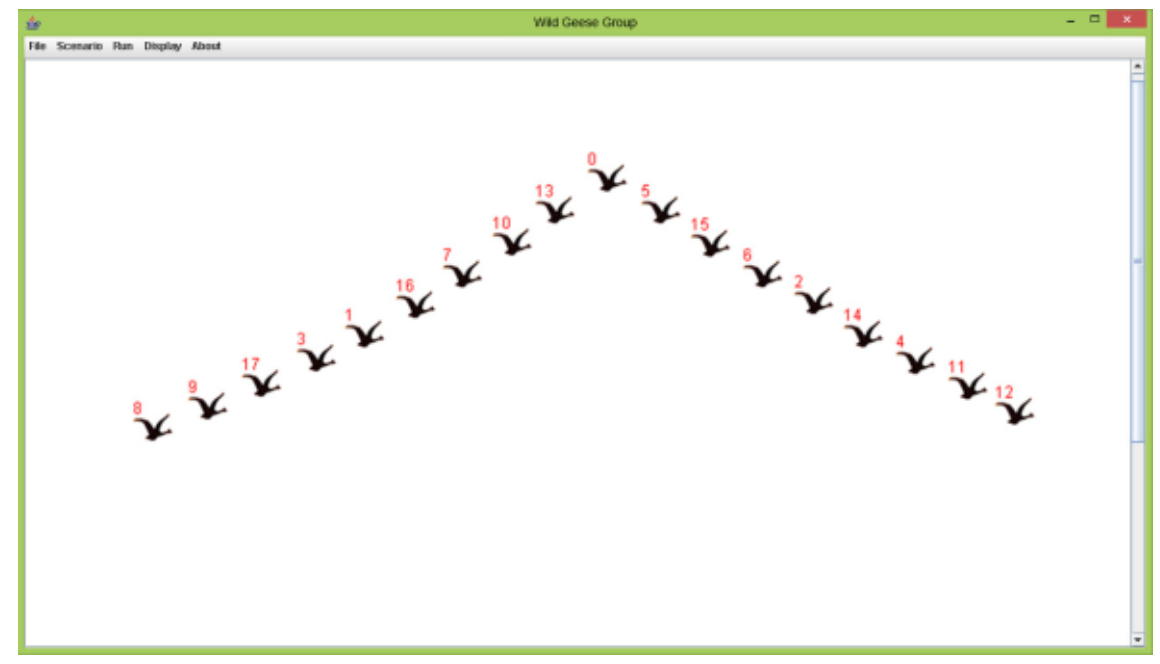

**Fig.4 Migrating Wild Geese**

In the migrating group of wild geese, the group size is 18, the least safe distance is supposed to be 60 cm (i.e., following becomes unsafe when the distance is less than 60 cm), and the biggest view angle is 128° (i.e., a goose cannot see others when they beyond –64° to 64°). In the beginning, the formation of the group is disordered. After a period of running, a distinct flying formation appears (Fig.4). When the leading goose is tired and slows down, another goose will take the place to lead the group after a while.

The swarm intelligence emerging from the migrating wild geese group contains two aspects. The first is that the group of geese form a distinct organization in which there is only one leading goose; the second is that the group gradually forms a distinct behavior pattern, which is a leading-goose-centric herringbone pattern. These aspects emerge as the geese are trying to maintain the balances of their contradictions.

## 5.4 Swimming Pool

In a swimming pool (for imitating a fish school's hedging), there are two kinds of swimmers, i.e., learners and veterans. It is common sense that veterans swim much faster than learners. Swimmers try their best to swim speedily and safely. Swimmers always keep a safe distance from other swimmers and the pool's sidewalls, in addition to not hindering others' swimming. A swimmer in the pool contains two contradictions:

- $c_{41}$. Distance safety *vs* Collision Dangerousness: is the following distance between the swimmer and the preceding swimmer (or the pool's sidewalls) big and safe enough so that the swimmer can swim at the preferred speed?
- $c_{42}$. Feeling uncrowdedness *vs* Crowdedness: are there so many swimmers in the pool that the swimmer cannot keep the normal swimming speed?

When swimmers are too close (i.e., the distance is unsafe and a collision is likely) to their preceding swimmers or the pool's sidewalls, they need to turn away to avoid running into others or hitting the walls. Furthermore, when swimmers feel that they could not swim at the preferred speed (e.g., because of crowding), veterans had better follow the tides of other veterans in order to maintain their preferred swimming speed and not collide into learners. Alternatively, learners should not swim into groups of veterans to avoid being hit by the fast-swimming veterans. Thus, for veterans, maintaining a fast enough swimming speed is more important; they try to keep swimming normally on the premise that they can swim safely. For learners, guaranteeing the safety is more significant in addition to not hindering the swimming of the veterans.

In the swimming pool, the size of the pool is 50 m × 50 m. There are 1000 swimmers among which 3/4 are learners (grey blocks in Fig.5) and 1/4 are veterans (red blocks). The normal swimming speed of a learner is 1 m/s whilst the speed of a veteran is 2 m/s. A swimmer feels crowded when the swimmer always cannot swim at the normal preferred speed. We set an empirical threshold for judging whether it is crowded, that is, a swimmer feels crowded when the number of surrounding swimmers the swimmer may run into at one stroke is over $\pi \times speed^2$. Moreover, we assume that veterans turn left (i.e., anticlockwise) by default when they are approaching the pool's sidewalls.

At the beginning, swimmers are scattered in the pool randomly and the pool is chaotic. After a period of time of running, a distinct swimming loop formed by veterans appears and learners gather into several clusters away from the veteran swimming loop (Fig.5).

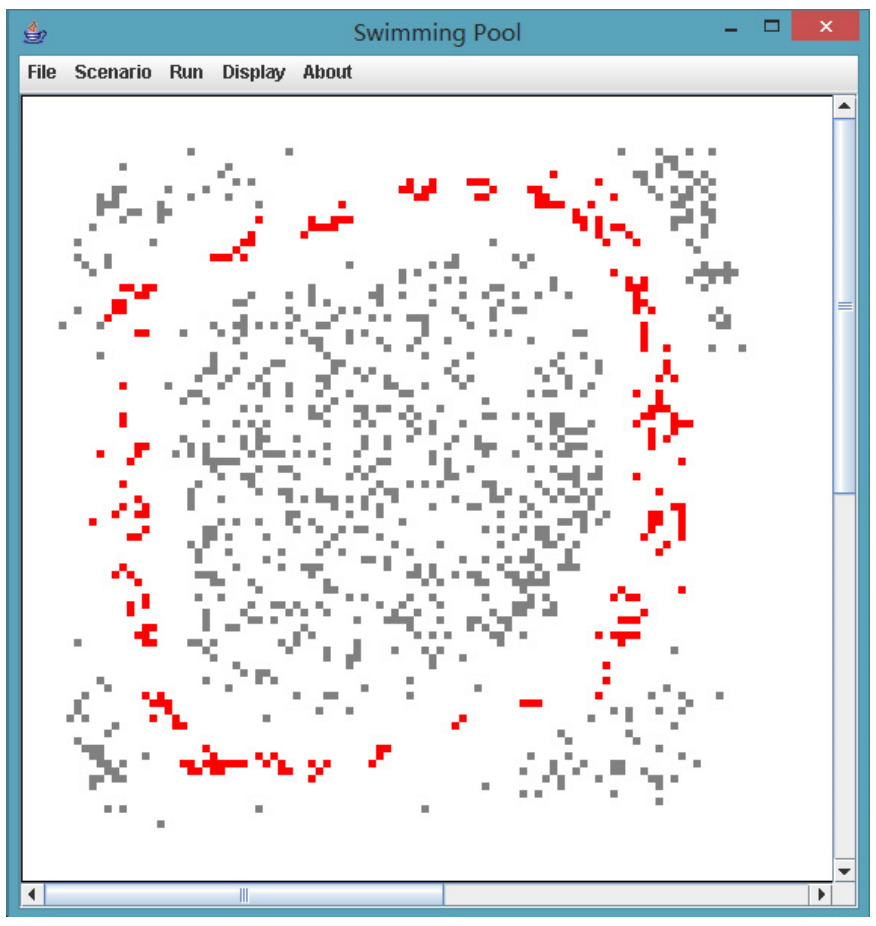

**Fig.5 Swimming Pool**

The swarm intelligence emerging from the swimming pool manifests as a distinct behavior pattern, i.e., a swimming loop. In the loop, veterans can always keep their swimming speeds; learners keep safe distances away from the veterans and are divided into several clusters located away from the swimming loop.

## 5.5 Market of an Economic Society

In a market, people have to earn their living. In other words, the daily activities of people in a market are to earn money to support themselves (including investing, e.g., setting up factories to produce consumer goods, or finding jobs to earn wages to purchase goods). Among people, some are adventurous whilst others tend to be conservative. Adventurous people are willing to invest by using their spare money, e.g., setting up factories to employ workers to produce and market products to make profits. Nevertheless, people must work to earn money to meet their daily needs when they are impoverished (i.e., they do not possess enough consumer goods or have adequate money for purchasing goods), no matter whether they are adventurous or conservative.

A person in a market contains two contradictions:

- $c_{51}$. Richness *vs* Poverty: Does the person have abundant assets (including money and consumer goods) to support consumption?
- $c_{52}$. Adventurousness *vs* Conservativeness: Does the person tend to invest to earn more profits when the person is affluent or still find a job that can earn a guaranteed wage?

First of all, people need to consume to survive. Investors (who must also be employer) can consume goods produced by employees (or workers); whilst employees without investments purchases goods for consumption. Second, adventurous persons invest to earn profits when they have a surplus over their consumption needs. After investors set up factories and employ workers to produce goods, they must put the residual goods (excluding those they consume) onto the market to realize profits. Third, conservative or poor people have to work to earn wages. After earning wages, they must go to the market to purchase goods since they cannot produce and consume goods like investors. It is easy to understand that every person always tries hard to accumulate wealth, i.e., to keep the relative strength of the contradiction of $c_{51}$ positive.

At the market level, every person consumes in the market. Meanwhile, an investor always tries to employ workers to produce goods and sell the goods to workers; whilst a worker always tries to find a job to be employed by some investor to earn a wage for purchasing goods. Thus, an investor is always associated with a group of workers whilst a worker can only be connected to a specific investor (provided that it is full employment). When consuming, a person becomes less rich; when employing workers, an investor can realize and possess the profits from the goods. When working, a worker can provide his labor to produce goods and earn a wage in return.

In the market, the population size is 1000, the initial assets that people possess are allotted either equally or randomly, and the market profit is set to be 10%. In order to observe the impact of the number of investors on the distribution of social wealth, the proportions of adventurous people are set at 30~80%.

We count the percentage of social wealth that the top 20% of the population possess (as shown in Fig.6). As shown in the statistical results, no matter whether the assets are allotted equally or randomly at the beginning, the top 20% of the population eventually acquire approximately 80% of the social wealth and the inequality of wealth continues afterward.

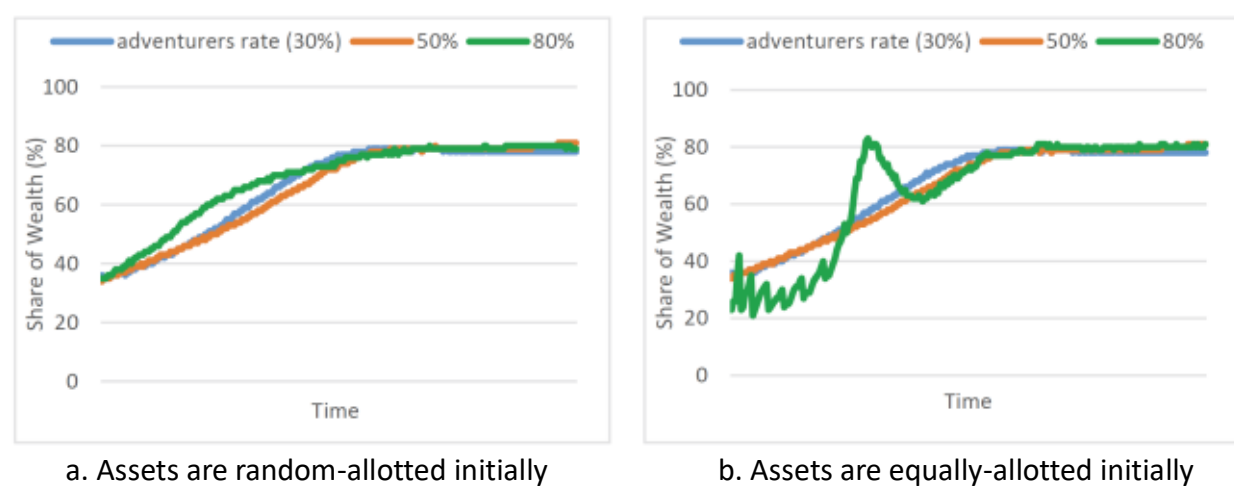

a. Assets are random-allotted initially    b. Assets are equally-allotted initially

**Fig.6 Social Wealth Share (top 20%)**

When assets are randomly allotted at the beginning, the fewer workers (who are conservative) there are,

the more quickly the inequality of wealth appears (Fig.6.a). Among the statistical results, there arises a curious phenomenon (Fig.6.b). When the assets are allotted equally at the beginning and there are more than 80% of the people are willing to invest, a transient extreme inequality of wealth occurs before the inequality becomes stable. This may be because some unlucky investors run out of their money simultaneously at that time due to too few workers in the market and their extreme poverty results in a temporary extreme inequality of wealth. After that, they had to work to earn money to support themselves and the inequality is relieved slightly. Nevertheless, a stable inequality eventually emerges and continues.

The continuous inequality of social wealth among people in the market reflects the 80/20 rule (or Pareto's principle). Accordingly, the swarm intelligence emerging from the market contains two aspects. The first is that the market forms a distinct organization in which people are either investors or workers, but all of them are consumers; the second aspect is that a distinct behavior pattern emerges from the market eventually, i.e., the 80/20 rule, and it subsequently remains steady.

# 6 Validations

We validate our model with two studies. In the first study, we conduct experiments to observe and analyze whether our model can properly explain the emergence of swarm intelligence. In the second study, we compare our approach with representative existing approaches to demonstrate the practicality of our model by observing the efficiencies of the evolution (or emergence) of swarm intelligence.

## 6.1 The Validity of Our Model

Generally speaking, a swarm intelligence model is valid if it can be used to explain those emergence phenomena we observe, or in other words, swarms simulated by using the model exhibit the emergence properties we expect. To validate the model, we adjust the parameters of the systems to observe what phenomena emerge from the swarms.

In a foraging ant system, we modify the size of the ant colony to observe the formation of the foraging path. It can be seen from the statistical results (Fig.7) that the foraging path from the food source to the nest does not always occur when the colony has a smaller size. This happens because the secreted pheromone from the food source to the nest may have evaporated cleanly before it is touched and enhanced by a small number of ants. Nevertheless, when the colony size is big enough, there are greater chances for ants to encounter the pheromone, find the food source, and further enhance the pheromone; eventually, the foraging path appears between the food source and the nest.

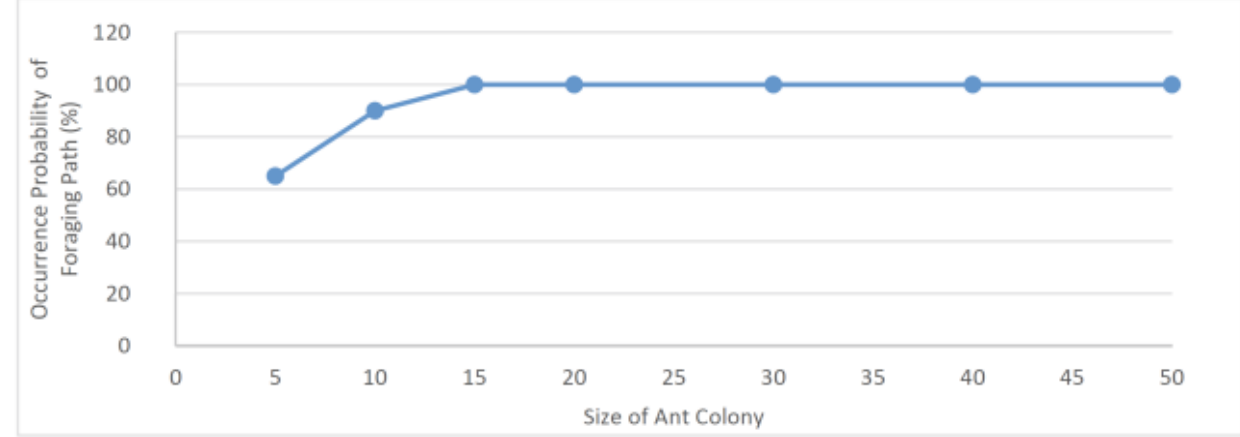

**Fig.7 Probability of Ant Foraging Path Occurrence**

In a bee colony, we observe the development of the queen bee when the swarm is of different sizes. As shown by the statistical results (Fig.8), there eventually is exactly one mature queen bee that survives in the colony when the swarm is small. However, when the size of the swarm becomes bigger (e.g., exceeding 10 or 22 thousand in our experiments), two or three matured queen bees survive in the swarm. In the real world, only one queen bee is allowed for each swarm, and the experimental results may explain why a swarm is split into sub-swarms when the swarm becomes larger. In fact, if several queen bees can coexist and the swarm is never split, our experiments show that only one queen bee ultimately survives in the colony.

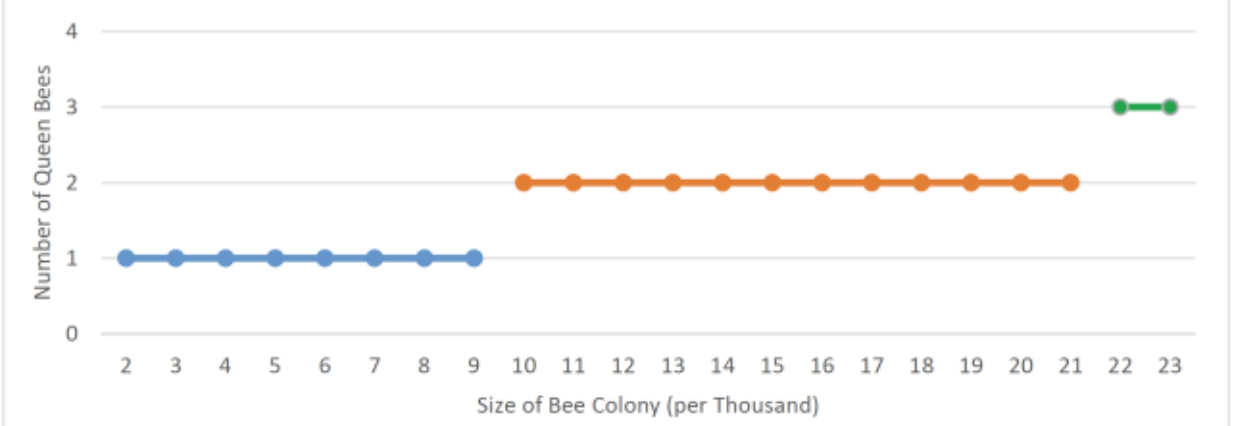

**Fig.8 Numbers of Queen Bees in Bee Colonies with Different Sizes**

In a group of wild geese, we change the group size to observe the flying formation. As the experimental results show (Fig.9), when the group size is small, a stable herringbone pattern appears; but when the group size becomes large (especially when the number of geese on one side exceeds 16), the flying formation cannot remain stable. When the group size is big, the geese at the end of the group are too far away from the leading goose, and they may determine that they cannot be well protected by the group. As a result, they speed up in order to keep up with the group and subsequently the flying formation cannot remain stable. Naturally, when a goose cannot join in a stable flying pattern, it is neither well-protected nor energy-saving and it eventually leaves the group. This also explains why the group size of the herringbone pattern we see in the real world is not very large.

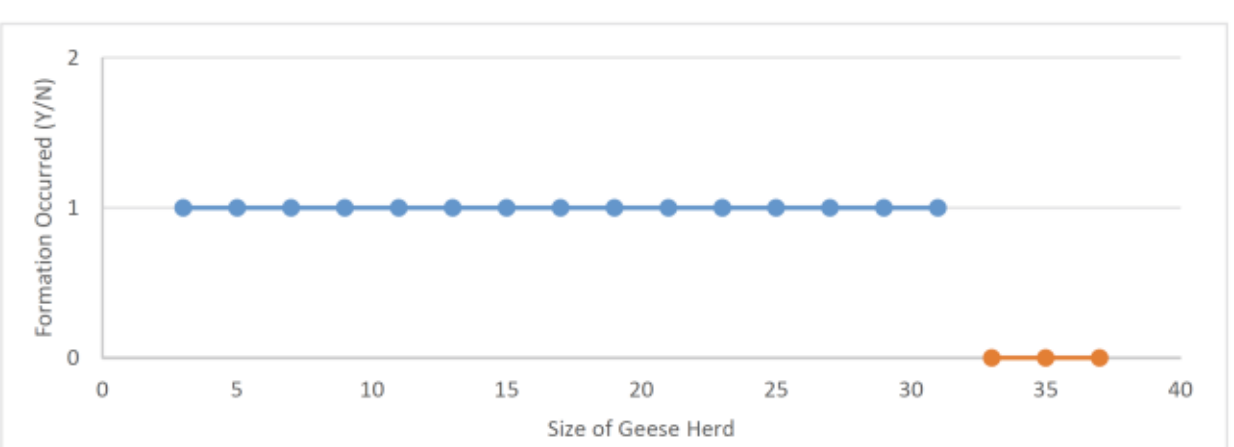

**Fig.9 The Occurrence of Geese Flying Formations**

In a swimming pool, we also change the number of swimmers in order to observe the behavior patterns of swimmers. As the statistical results show (Fig.10), when there are many swimmers in the pool and the pool becomes crowded, a swimming loop formed by veterans appears. However, when there are few swimmers and the pool is not crowded, veterans have enough space to swim at the normal speed and they do not necessarily swim in a loop.

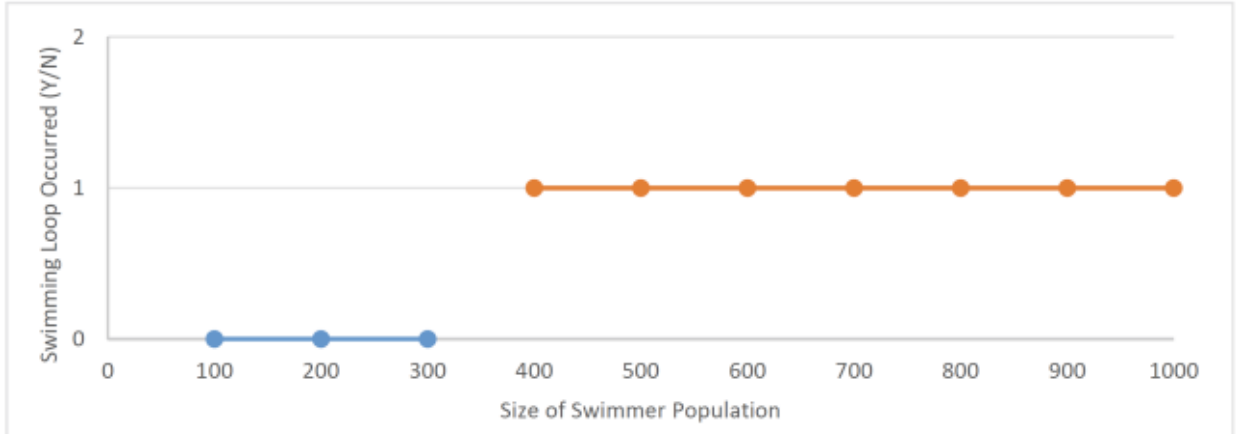

**Fig.10 The Occurrence of Swimming Loop**

In a market, we observe the distribution of social wealth by setting whether the investors need to find jobs as workers in addition to employing workers. As the experimental results show (Fig.11), there are two cases to consider. In the first case, when investors do not need to find jobs, i.e., they accumulate their wealth completely through investments, the social wealth becomes highly concentrated (i.e., 99% of the wealth is possessed by the top 20% of the population) when there are fewer investors (e.g., less than 20% of the population). The distribution of the social wealth is in compliance with the 80/20 rule when there are more investors (e.g., over 25% of the population). In the second case, when investors and workers need to find jobs, the distribution of the social wealth becomes more even. This trend is strengthened as the rate of investors in the population increases. Working is not only a means of creating wealth, but it is also a way of transferring wealth (from workers to investors). Therefore, if investors do not work, the transfer of social wealth is unidirectional and the possession of the social wealth becomes polarized (i.e., the 80/20 rule emerges). However, when investors also work to accumulate wealth, they are creating profits for other investors through their employment in addition to earning wages; hence, the social wealth becomes more evenly distributed.

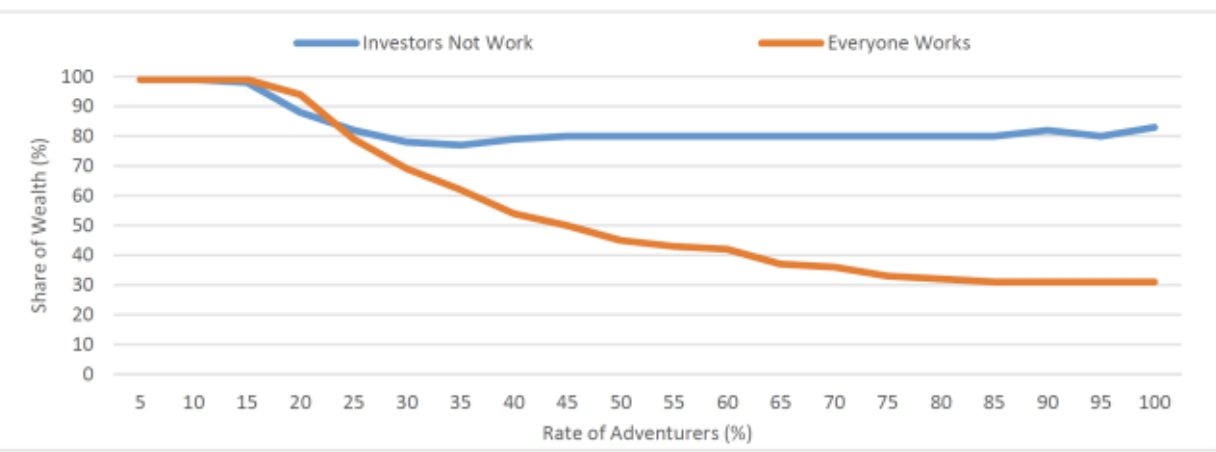

**Fig.11 Share of Wealth for Populations with Different Rates of Adventurer**

It can be seen from the experimental results that the systems can present emergence phenomena in compliance with our intuition, and it further indicates that our model is widely applicable to the emergence of swarm intelligence.

## 6.2 Efficiency Comparison

In our model, individuals decide and choose their behavior instinctively based on their internal contradictions, so they do not need any complex deliberation or computation. In this sub-section, we compare the time complexity of our model with representative approaches (e.g., ACO [11] and PSO [8]) from the aspect of the evolution of swarm intelligence. We implement ACO and PSO in the ants foraging system and the wild geese system, respectively, and compare the time costs (Fig.12).

The implementation platform is with a configuration with an Intel ® Core™ i7-7660U CPU @ 2.50 GHz and 8.00 GB RAM. We run the systems 10 times and calculate the average time costs.

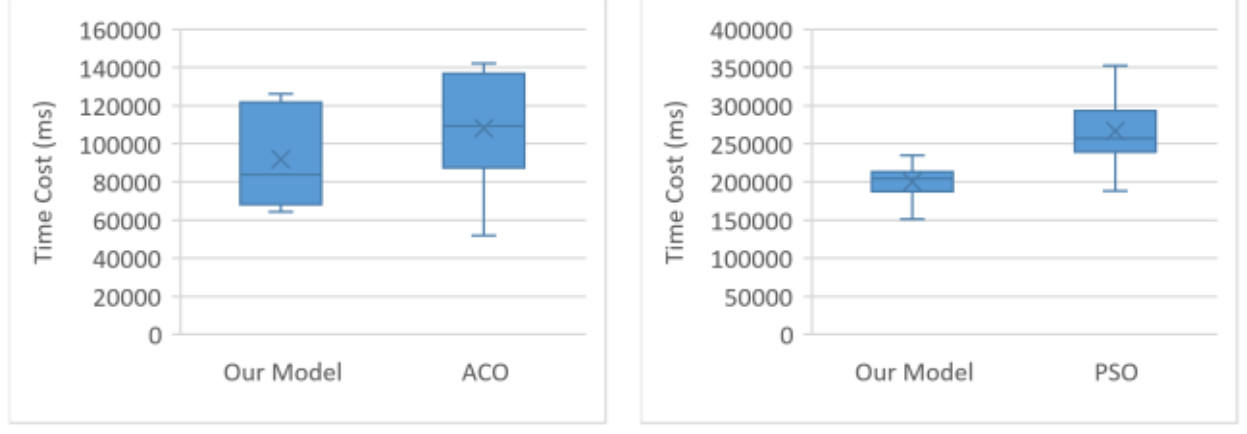

**Fig.12 Time Cost Comparisons between our Model and Existing Approaches**

As shown in Fig.12.a, the foraging path forms somewhat faster when the system is implemented by using our model than with the ACO algorithm. In the ACO implementation, ants do not secrete pheromones until they carry and unload the food to the nest, which postpones the emergence of the foraging path. In Fig.12.b, the group of geese can be well-formed much more quickly by using our model than with the PSO algorithm. When implemented in the PSO algorithm, a goose observes the behavior and positions of the other geese surrounding it and then decides its own behavior. The PSO based approach requires more computations and consequently incurs a higher time overhead.

### 6.3 Practicality Analysis

In the related work section (Section 3), we analyze and describe the application scopes and limitations of representative existing approaches. Table 2 displays the applicability of the representative approaches to the systems simulated previously (see Section 5).

**Table 2. Applicability of Swarm Intelligence Models**

| Model | Foraging Ants | Queen Bee | Geese Group | Swimming Pool | Market |
|---|---|---|---|---|---|
| CA | √ | | √ | √ | |
| ACO | √ | | | √ | |
| PSO | √ | | √ | √ | |
| ABC | √ | | | √ | |
| Our Model | √ | √ | √ | √ | √ |

In combination with the results of the above experiments, we can assert that our model is more generic to study the emergence of swarm intelligence.

## 7 Conclusions and Discussions

Determining the root causes for the emergence of swarm intelligence remains a key challenge. Existing studies are either they are too abstract and difficult to apply to practical problems or they are too specific and only applicable to specific swarm intelligence problems. To address this issue, the essential root cause of the emergence of swarm intelligence is investigated in this work. A novel model is proposed that is based in the development of contradictions within interacting individuals.

As far as we know, our proposed model is the most truly generic and straightforward model for swarm intelligence. As illustrated by the case studies, our model can describe the emergence of various types of swarm intelligence. Among the systems, some have the emergence of a holistic structure (e.g., queening bee), some have the emergence of a global behavior mode (e.g., foraging ants and swimming pool), and some have both (e.g., migrating wild geese and markets). Since the key to modeling a swarm is to find the contradictions inside individuals and the relevances between individuals and the environment, it is also very simple and easy to implement the emergence of swarm intelligence without needing complex computations.

Besides implementing five systems to illustrate the generality of our model, we also do some experiments to verify its validity and practicality. By re-implementing the systems with varied parameter configurations, we can find that the phenomena emerging from the systems implemented by using our model are in compliance with our intuition, which indicates that our model is valid to model swarm intelligence in the real world. In addition, by comparing the proposed model with other existing approaches, we find that our model is more efficient to implement swarm intelligence as it requires fewer computations.

However, there are still some limitations in the current model that need additional investigation in the future. Firstly, the contradictions are predetermined and the domination relationships between contradictions and appearances are also pre-specified and fixed. Currently, new contradictions can arise while individuals are interacting, but for simplicity, while formalizing interactions among individuals, we have assumed that the newly arisen contradictions are foreseeable so that we can specify the domination relationships between contradictions and appearances in advance.

We believe that the expected emerging result of swarm intelligence can be ensured if we can appropriately specify contradictions and interactions. Nevertheless, it may not be so easy to discover the correct and underlying contradictions inside individuals, and even if we can specify the contradictions appropriately in advance and imagine what emergence phenomena can happen, we still cannot precisely predict the emergence results. For instance, in the queening bee system, a bee just interacts with part of the swarm instead of all of the other bees in the swarm and we cannot expect that there is only one queen to win out in the end. Similarly, in the market system, although we can imagine that some kind of inequality of wealth takes place eventually, we do not foresee the emergence of the 80/20 rule.

A second limitation is that the model only takes into consideration contradictions at the individual level, but it does not handle contradictions at the swarm level. There must be contradictions at the swarm level, e.g., social contradictions, although swarm level contradictions essentially emerge from individual contradictions.

A third limitation is that the model can be applied to explaining the emergence of swarm intelligence, but it cannot describe the evolution of swarm intelligence. To enable the evolution of swarm intelligence, the relevances and interactions among individuals should be dynamic so that new configurations and distributions of contradictions can emerge and evolve.

In the future, we plan to extend the model to enable dynamic conditions, for instance, new contradictions may arise, new appearances may take place, new individuals may be generated, and swarm intelligence may evolve while individuals are behaving and interacting in the swarm. Furthermore, the in-teractions specified in the model are coarsely described at present. We plan to establish a mathematical calculus system for individuals and their interactions so that the emergence of swarm intelligence can be inferred rigorously.

## Contribution of individual authors to the creation of a scientific article (ghostwriting policy)

Wenpin Jiao carried out the whole work solely.

## Sources of funding for research presented in a scientific article or scientific article itself

- National Science and Technology Major Project of China (2020AAA0109400)
- National Natural Science Foundation of China (61620106007, 62072007).